\documentclass[conference,a4paper]{IEEEtran}
\IEEEoverridecommandlockouts
\usepackage{cite}
\usepackage{amsmath,amssymb,amsfonts}
\usepackage{graphicx}
\usepackage{textcomp}
\usepackage{xcolor}
\usepackage{algorithm}
\usepackage{algpseudocode}
\usepackage{multirow}
\usepackage{listings}
\usepackage{mdframed}
\usepackage{caption}
\usepackage{float}
\usepackage{placeins}
\usepackage{dblfloatfix}
\usepackage[hyphens]{url}
\usepackage{cite}

\def\BibTeX{{\rm B\kern-.05em{\sc i\kern-.025em b}\kern-.08em
    T\kern-.1667em\lower.7ex\hbox{E}\kern-.125emX}}

\newcommand{\tightsection}[1]{%
  \vspace{-1.5em}%
  \section{#1}%
  \vspace{-0.5em}%
}

\begin{document}

\title{Reliable Text-to-SQL in Resource-Constrained Settings with Small LLMs via Iterative Retry Workflows}

\renewcommand{\IEEEauthorrefmark}[1]{\textsuperscript{#1}}
\author{
\IEEEauthorblockN{Bilel Abderrahmane Benziane\IEEEauthorrefmark{1} and Xavier Schepler\IEEEauthorrefmark{1}}
\IEEEauthorblockA{
\IEEEauthorrefmark{1}Data Lab, Generix Group, Paris, France\\
Emails: babenziane@generixgroup.com, xschepler@generixgroup.com
}
}

\maketitle

\makeatletter
\AddToHookNext{shipout/foreground}{%
  \put(35,-785){%
    \parbox{0.44\paperwidth}{%
      \scriptsize
      Published in: 2026 31st International Conference on\\
      Automation and Computing (ICAC)\\
      IEEE Xplore\\
      DOI: 10.1109/ICAC70325.2026.11712531
    }%
  }%
}
\makeatother

\begin{abstract}
Large Language Models (LLMs) have recently improved Text-to-SQL generation, but their computational cost can hinder deployment in resource-constrained settings. This paper evaluates the Text-to-SQL capabilities of small and medium open-source LLMs (1B--8B parameters) under a controlled inference budget, focusing on how model specialization and prompt formulation affect performance. Using the Spider benchmark, we compare multiple prompt representations and assess the accuracy of generated queries. Our results show that SQL-specialized models are consistently more reliable than general-purpose and chat-oriented models at small scales, and that performance can vary substantially depending on the prompt used. To reduce this variability without resorting to larger models, we introduce three lightweight inference workflows that automatically retry with alternative prompts and/or models when failures occur, including an optional syntax-correction step. These workflows significantly improve end-to-end success rates, with a trade-off in additional inference calls. These results demonstrate a practical path to stronger Text-to-SQL performance with small LLMs, by trading modest additional inference cost for higher reliability.
\end{abstract}


\tightsection{Introduction}
With the abundance of data in the last few years, artificial intelligence has witnessed rapid growth in the use of foundation models \cite{bommasani2022opportunitiesrisksfoundationmodels}. Training a model on a wide range of data can help improve generalization and offer better inference on unseen data. One AI subfield that has witnessed a rapid increase in performance is natural language processing, where the use of large language models (LLMs) has rapidly increased in the last few years~\cite{devlin2019bert,brown2020language}.

However, a major limitation of LLMs is their lack of knowledge on specific tasks. To increase the performance of an LLM, it has to be tuned on a specific type of data or field \cite{gururangan2020dontstoppretrainingadapt}. As a result of fine-tuning LLMs, many specialized models have emerged in the last few years, such as instruction-tuned/chat LLMs, code LLMs, and domain-specific LLMs~\cite{wei2022finetuned,ouyang2022training,chen2021evaluating,wu2023bloomberggpt,luo2022biogpt}. One special type of code LLM is Text-to-SQL. The importance of these specialized models specifically lies in their direct impact on democratizing data access by offering a means for people with a non-technical background to query complex databases.

Although LLMs are important in facilitating tasks, they tend to be resource-hungry. Unlike traditional smaller models, LLMs require much more expensive resources \cite{kaplan2020scaling}. This could lead to an exponential increase in the business cost of any given software solution making use of an LLM. Previous work tackled this problem by compressing the models to lower numerical precision. One example is the Q8BERT model, an 8-bit quantized version of BERT obtained via quantization-aware fine-tuning, achieving roughly 4$\times$ compression and potential inference speedups with negligible accuracy loss \cite{zafrir2019q8bert}. While these methods reduce memory use, they degrade performance, usually require tuning quantization further after training, and have only been studied for models with fewer than 350M parameters.

Previous research tackled the same problem on a larger scale by carrying out more extensive experimentation on large and medium-sized LLMs \cite{gao2023texttosqlempoweredlargelanguage}. The results showcased a positive correlation between model scale and performance, where the larger version of fine-tuned LLaMA-2 (CodeLLaMA-34B) achieved an average execution accuracy of 39.6\% compared to the smaller version (LLaMA-2-CHAT-13B), which achieved an average execution accuracy of 19.4\%. This work also showed that question representation plays an important role in improving performance. Building on previous work, this study focuses on assessing the performance of medium and small LLMs for the Text-to-SQL task. In order to improve the performance of smaller LLMs, this work suggests three different workflows to improve accuracy. 

The remainder of this paper is organized as follows. Section~\ref{sec:related-work} presents the related work. Section~\ref{sec:methodology} describes the proposed methodology. Section~\ref{sec:results} reports and discusses the experimental results. Finally, Section~\ref{sec:conclusion} concludes this work and outlines possible directions for future research.

\section{Related Work}\label{sec:related-work}
Many paths exist in the literature to improve the performance of LLMs. In the next sections, two main techniques are discussed.

\subsection{Fine Tuning}
Fine-tuning a generic LLM is the first step toward obtaining a specialized Text-to-SQL model. For instance, Qwen2.5-Coder-7B-Instruct results from training Qwen2.5 on a large corpus of programming languages and not only SQL \cite{hui2024qwen25coder}. Deepseek-ai/deepseek-coder-6.7b-instruct is a tuned version of deepseek-ai/deepseek-coder-6.7b-base \cite{guo2024deepseekcoder}. Similarly, \texttt{codellama/CodeLlama-7b-Instruct-hf} is the instruction-tuned variant of Code Llama 7B, a code-specialized model obtained by further training Llama 2 on code-centric data and then aligning it for instruction following \cite{roziere2023codellama}.

These models have between 6 and 8 billion parameters, which is considered medium-sized. Smaller models with as few as 1 billion parameters have appeared recently. These small models are not only code-specialized but also SQL-specialized. One example is \texttt{seeklhy/codes-1b-spider}, which is fine-tuned using a curated SQL-centric corpus \cite{li2024codes}. Another recently released SQL generation model is \texttt{prem-research/prem-1B-SQL}. This model is based on DeepSeek-Coder-1.3B-Instruct \cite{guo2024deepseekcoder}. Due to the abundance of open-sourc fine-tuneded code models, this work focuses on evaluating already pre-trained models.

\subsection{Prompt Styles}
Another part of the literature focuses on prompt engineering for the query generation process. Any given question can be enriched with extra context before being passed to the model to ensure that it has enough information for SQL generation. This process is known as question representation. The most basic type of question representation is called the Basic Prompt \cite{pourreza2023dinsqldecomposedincontextlearning}. This representation includes each table name and columns, followed by the prefix ``Q:'' and the original natural language question, and an answer prefix such as ``A: SELECT'' to indicate that the output should be an SQL query.

Another type of prompt is the Text Representation \cite{nan2023enhancingfewshottexttosqlcapabilities}. This prompt is similar to the Basic Prompt, with an additional instruction at the beginning to make the task clearer. OpenAI released their own prompt for this task, which further clarifies the format by adding the symbol ``\#'' to information fields and explicitly asking the model to complete the SQL query without explanation \cite{liu2023comprehensiveevaluationchatgptszeroshot}. Another suggested prompt is Alpaca SFT \cite{taori2023stanfordalpaca}. This prompt is very similar to the OpenAI prompt, but the instruction is more detailed, as it was originally designed for fine-tuning purposes.

A common limitation of the aforementioned prompting styles is the lack of full table schemas, which is crucial for generating SQL queries that join multiple tables. For that purpose, the Core Representation Prompt adds the full SQL schema for each table to the question \cite{chang2023promptllmstexttosqlstudy}. This includes foreign keys, which are important for generating multi-table queries.

Despite the abundance of papers on LLMs, evaluations of medium and small LLMs for SQL generation remain scarce. To fill this research gap, this work focuses on evaluating multiple well-established LLMs for the Text-to-SQL task across different prompting styles and SQL query types. In order to improve the performance of small LLMs, three SQL generation workflows are suggested and evaluated against a baseline.

\section{Methodology}\label{sec:methodology}
This section presents the experimental setup used in this work. We describe the dataset, the evaluation metrics, the set of models considered, the prompting configurations used for inference, and the proposed workflows designed to improve robustness in resource-constrained settings.

\subsection{Dataset}
\begin{figure}[htbp]
    \centering
    \includegraphics[width=0.30\textwidth]{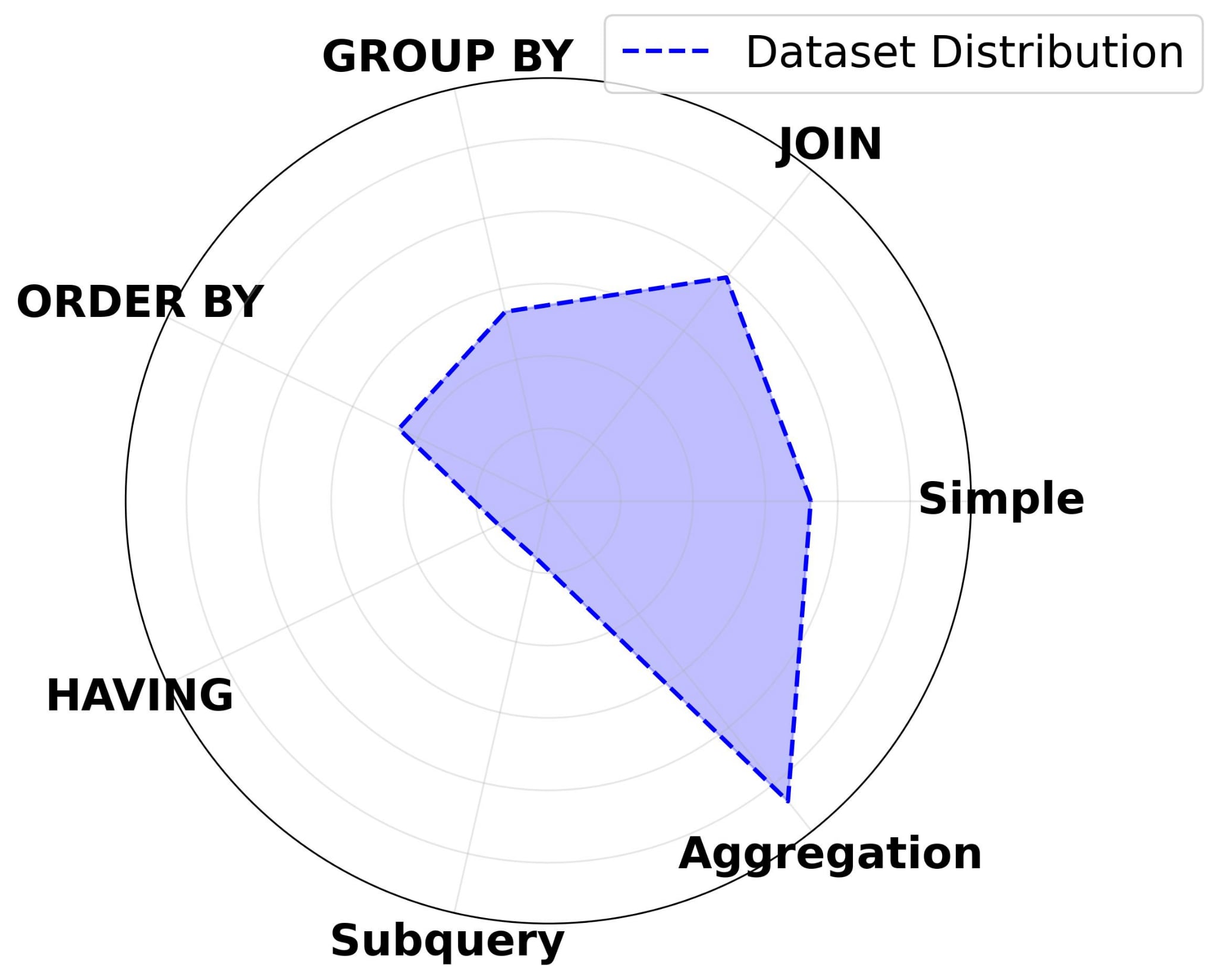}
    \caption{Distribution of SQL query types in the Spider evaluation set.}
    \label{fig:spider-sql-feature-distribution}
\end{figure}

For evaluation, we use the Spider dataset \cite{yu2019spiderlargescalehumanlabeleddataset}. Spider contains 10,181 cross-domain natural language questions collected from 200 databases, paired with corresponding SQL queries. Following the experimental setup suggested by \cite{gao2023texttosqlempoweredlargelanguage}, we use the official Spider development partition, comprising the last 1,034 examples, as the evaluation set. Figure~\ref{fig:spider-sql-feature-distribution} illustrates the distribution of query types in the evaluation data: simple SELECT, GROUP BY, ORDER BY, JOIN, AGGREGATE, SUBQUERY, and HAVING.

\subsection{Metrics}
To make a fair comparison, we follow the evaluation protocol used in prior work \cite{gao2023texttosqlempoweredlargelanguage}. To evaluate performance, we use exact-set-match accuracy (EM) and execution accuracy (EX). EM checks whether the predicted SQL query exactly matches the ground-truth query at the clause level (i.e., set-level comparison of keywords and predicates), making it a conservative metric: two semantically equivalent queries that differ in form will not match. EX, by contrast, compares the execution results of the predicted and ground-truth queries on the target database. Because a single question can often be answered by multiple semantically equivalent SQL queries, EX typically provides a more faithful estimate of model performance.

\subsection{Models}
To study how model specialization affects Text-to-SQL performance, we evaluate models from multiple families. Specifically, the tested models are grouped into five categories: (i) \emph{fundamental} (base) LLMs, (ii) \emph{instruction-tuned} LLMs, (iii) \emph{code} LLMs, (iv) \emph{chat} LLMs, and (v) \emph{SQL-specialized} LLMs.

\textbf{Fundamental models.} We evaluate base checkpoints including Llama-3-8B \cite{grattafiori2024llama3}, Qwen2.5-7B \cite{qwen2024qwen25}, Mistral-7B \cite{jiang2023mistral7b}, DeepSeek-LLM-7B-base \cite{deepseek2024deepseekllm}, and TinyLlama-1.1B \cite{zhang2024tinyllama}.

\textbf{Instruction-tuned models.} We also test instruction-tuned variants, including Llama-3-8B-Instruct \cite{grattafiori2024llama3}, Qwen2.5-7B-Instruct \cite{qwen2024qwen25}, and Mistral-7B-Instruct \cite{jiang2023mistral7b}.

\textbf{Code models.} For code-oriented generation, we evaluate Qwen2.5-Coder-7B and its instruction-tuned variant \cite{hui2024qwen25coder}, CodeLlama-7B and CodeLlama-7B-Instruct \cite{roziere2023codellama}, DeepSeek-Coder-6.7B base and instruct \cite{guo2024deepseekcoder}, StarCoder2-3B \cite{lozhkov2024starcoder2}, and CodeS-1B \cite{li2024codes}.

\textbf{Chat models.} To represent chat-optimized checkpoints, we include DeepSeek-LLM-7B-Chat \cite{deepseek2024deepseekllm}, Vicuna-7B v1.5 \cite{chiang2023vicuna}, Llama-2-7B-Chat \cite{touvron2023llama2}, and TinyLlama-1.1B-Chat \cite{zhang2024tinyllama}.

\textbf{SQL-specialized models.} Finally, we test models explicitly targeted at Text-to-SQL, including Defog-SQLCoder-7B \cite{defog2024sqlcoder7b2}, Prem-Text-to-SQL-1B \cite{premai2024prem1bsql}, and CodeS variants fine-tuned for Text-to-SQL, including CodeS-1B-spider and CodeS-1B-bird \cite{li2024codes}.

To ensure a fair comparison across methods, we allocate 150 tokens for response generation for all models. By default, we set the temperature to 0 to eliminate the influence of randomness. Each model is tested using the prompt styles discussed in Section~\ref{sec:related-work}. In addition, we introduce a \emph{structured prompt} (Fig.~\ref{fig:prompt-grid}) that organizes the input into explicit sections (instruction, database schema, and question). This prompt follows a structured format while using a simplified schema description, enabling a more consistent comparison across prompt styles.

\makeatletter
\@ifundefined{ptitle}{%
  \newcommand{\ptitle}[1]{\textsf{\scriptsize\textbf{#1}}}%
}{}

\@ifundefined{promptbox}{%
  \newmdenv[
    linewidth=0.5pt,
    innerleftmargin=5pt,
    innerrightmargin=5pt,
    innertopmargin=4pt,
    innerbottommargin=4pt,
    skipabove=3pt,
    skipbelow=3pt
  ]{promptbox}%
}{}
\makeatother

\begingroup
\lstset{
  basicstyle=\ttfamily\scriptsize\linespread{0.85}\selectfont,
  frame=none,
  breaklines=true,
  columns=fullflexible,
  keepspaces=true,
  showstringspaces=false,
  numbers=none,
  aboveskip=1pt,
  belowskip=1pt,
  lineskip=-1pt
}

\begin{figure}[tb]

\centering

\begin{promptbox}
\ptitle{Structured Prompt}\par\vspace{1pt}
\begin{lstlisting}
### Instruction:
Translate the following question into SQL.

### Database Schema:
continents (ContId, Continent)
countries (CountryId, CountryName, Continent)

### Question:
How many continents are there ?

### SQL:
SELECT
\end{lstlisting}
\end{promptbox}
\vspace{-5pt}
\caption{Structured prompt template for Text-to-SQL inference.}
\label{fig:prompt-grid}
\end{figure}

\endgroup
\subsection{Proposed Workflows}\label{sec:workflows}
Model performance may vary with both the question type and the prompt representation. As a result, using a single model and a single prompt style in an inference system can lead to a higher number of errors. To improve robustness, we propose three workflows.
\textbf{Single-Model/Prompt Baseline.}
This algorithm defines the baseline setting in which both the prompt style $P$ and the model $M$ are fixed. Given a question $q$, the system constructs a prompt $p$ using $P$, generates an SQL query $\hat{s}$ using $M$, and returns the execution result $R$ if $\hat{s}$ executes successfully.

\subsubsection{Prompt Workflow}
In the prompt workflow (Algorithm~\ref{alg:prompt_flow_text_to_sql}), the model $M$ is fixed and the system iterates over a list of prompt styles $P = [P_1,\dots,P_K]$. For each $P_i$, the system generates a candidate query $\hat{s}_i$ and attempts execution. If execution fails due to an SQL validation error, the same model $M$ is used once to produce a corrected query $\tilde{s}_i$ conditioned on $(q,\hat{s}_i)$. The workflow returns the first result $R$ obtained from a valid query; otherwise, it proceeds to $P_{i+1}$ until all prompt styles are exhausted.

\begin{algorithm}[!h]
\caption{Prompt-Flow with Syntax Correction}
\label{alg:prompt_flow_text_to_sql}
\begin{algorithmic}[1]
\scriptsize
    \State \textbf{Input:} Question $q$, list of prompt styles $P = [P_1,\dots,P_K]$,  model $M$
    \State \textbf{Output:} Query result set $R$ or failure
    \State Define prompt styles list $P$
    \State Define Text-to-SQL model $M$
    \State Read question $q$
    \State $i \gets 1$
    \While{$i \le K$}
        \State Format $q$ using prompt style $P_i$ to obtain prompt $p_i$
        \State Pass $p_i$ to model $M$ to generate SQL query $\hat{s}_i$
        \State Execute $\hat{s}_i$ on the database
        \If{no SQL validation error}
            \State Obtain result $R$
            \State \Return $R$
        \Else
            \State Pass $\hat{s}_i$ and $q$ to $M$ for syntax correction to obtain $\tilde{s}_i$
            \State Execute $\tilde{s}_i$ on the database
            \If{no SQL validation error}
                \State Obtain result $R$
                \State \Return $R$
            \Else
                \State $i \gets i + 1$
            \EndIf
        \EndIf
    \EndWhile
    \State \Return failure  \Comment{All prompt styles exhausted with invalid SQL}
\end{algorithmic}
\end{algorithm}

\subsubsection{Model Workflow}
In the model workflow (Algorithm~\ref{alg:model_flow_text_to_sql}), the prompt style $P$ is fixed and the system iterates over a list of models $M = [M_1,\dots,M_K]$. For each $M_i$, the system generates a candidate query $\hat{s}_i$ from the same prompt $p$ and attempts execution. If execution fails due to an SQL validation error, the same model $M_i$ is used once to generate a corrected query $\tilde{s}_i$ conditioned on $(q,\hat{s}_i)$. The workflow returns the first result $R$ obtained from a valid query; otherwise, it proceeds to the next model.

\begin{algorithm}[!ht]
\caption{Model-Flow with Syntax Correction}
\label{alg:model_flow_text_to_sql}
\begin{algorithmic}[1]
\scriptsize
    \State \textbf{Input:} Question $q$, prompt style $P$, list of models $M = [M_1,\dots,M_K]$
    \State \textbf{Output:} Query result set $R$ or failure
    \State Define prompt style $P$
    \State Define Text-to-SQL models list $M$
    \State Read question $q$
    \State $i \gets 1$
    \While{$i \le K$}
        \State Format $q$ using prompt style $P$ to obtain prompt $p$
        \State Use model $M_i$ to generate SQL query $\hat{s}_i$ from $p$
        \State Execute $\hat{s}_i$ on the database
        \If{no SQL validation error}
            \State Obtain result $R$
            \State \Return $R$
        \Else
            \Comment{Syntax correction by the same model}
            \State Pass $\hat{s}_i$ and $q$ to $M_i$ to obtain corrected SQL $\tilde{s}_i$
            \State Execute $\tilde{s}_i$ on the database
            \If{no SQL validation error}
                \State Obtain result $R$
                \State \Return $R$
            \Else
                \State $i \gets i + 1$  \Comment{Try next model in $M$}
            \EndIf
        \EndIf
    \EndWhile
    \State \Return failure \Comment{All models used and SQL remains invalid}
\end{algorithmic}
\end{algorithm}

\subsubsection{Prompt-and-Model Workflow}
In the prompt-and-model workflow (Algorithm~\ref{alg:model_prompt_flow}), the system iterates over combinations of prompt styles and models. Given prompt styles $P = [P_1,\dots,P_K]$ and models $M = [M_1,\dots,M_L]$, the workflow loops over models $M_j$ and, for each $M_j$, loops over prompt styles $P_i$. For each pair $(M_j,P_i)$, the system generates $\hat{s}_{i,j}$, attempts execution, and, upon validation failure, applies a single correction step using the same model $M_j$ to produce $\tilde{s}_{i,j}$ conditioned on $(q,\hat{s}_{i,j})$. The workflow returns the first valid result $R$; if all pairs fail, it returns failure.

\begin{algorithm}[!ht]
\scriptsize
\caption{Model-and-Prompt Flow with Syntax Correction}
\label{alg:model_prompt_flow}
\begin{algorithmic}[1]

    \State \textbf{Input:} Question $q$, prompt styles list $P = [P_1,\dots,P_K]$, models list $M = [M_1,\dots,M_L]$
    \State \textbf{Output:} Query result set $R$ or failure
    \State Define prompt styles list $P$
    \State Define Text-to-SQL models list $M$
    \State Read question $q$
    \State $j \gets 1$  \Comment{model index}
    \While{$j \le L$}
        \State $i \gets 1$  \Comment{prompt-style index}
        \While{$i \le K$}
            \State Format $q$ using prompt style $P_i$ to obtain prompt $p_{i,j}$
            \State Use model $M_j$ to generate SQL query $\hat{s}_{i,j}$ from $p_{i,j}$
            \State Execute $\hat{s}_{i,j}$ on the database
            \If{no SQL validation error}
                \State Obtain result $R$
                \State \Return $R$
            \Else
                \Comment{Syntax correction with same model}
                \State Pass $(q,\hat{s}_{i,j})$ to $M_j$ for syntax correction to obtain $\tilde{s}_{i,j}$
                \State Execute $\tilde{s}_{i,j}$ on the database
                \If{no SQL validation error}
                    \State Obtain result $R$
                    \State \Return $R$
                \Else
                    \State $i \gets i + 1$  \Comment{Try next prompt style}
                \EndIf
            \EndIf
        \EndWhile
        \State $j \gets j + 1$  \Comment{All prompts tried, move to next model}
    \EndWhile
    \State \Return failure \Comment{All $(M_j,P_i)$ combinations failed}
\end{algorithmic}
\end{algorithm}

\section{Results and Discussion}\label{sec:results}
This section reports the performance of the evaluated models across the different prompt styles. Table~\ref{tab:prompt_style_results} summarizes execution accuracy (EX) and exact-set-match accuracy (EM) for each model and prompt style.

\begin{table*}[h!]
\centering
\tiny
\resizebox{\textwidth}{!}{%
\begin{tabular}{llcccccccccccccc}
\hline
Type & Model ID & \multicolumn{2}{c}{\shortstack{Alpaca\\ Sft}} & \multicolumn{2}{c}{\shortstack{Basic\\ Prompt}} & \multicolumn{2}{c}{\shortstack{Code\\ Repr.}} & \multicolumn{2}{c}{\shortstack{OpenAI\\ Demo}} & \multicolumn{2}{c}{\shortstack{Struct.\\ Prompt}} & \multicolumn{2}{c}{\shortstack{Text\\ Repr.}} & \multicolumn{2}{c}{Avg} \\
 &  & EM & EX & EM & EX & EM & EX & EM & EX & EM & EX & EM & EX & EM & EX \\
\hline
\multirow{6}{*}{Fund.} & Llama-3-8B & 21.86 & 70.50 & 2.32 & 33.08 & 28.53 & 3.10 & 29.21 & 6.29 & 25.15 & 74.95 & 25.05 & 39.27 & 22.02 & 37.86\\
 & Qwen2.5-7B & 31.43 & 87.81 & 4.45 & 27.66 & 39.27 & 43.42 & 37.91 & 30.37 & 37.23 & 83.08 & 37.91 & 90.43 & 31.37 & 60.46\\
 & Mistral-7B & 21.95 & 27.27 & 2.03 & 27.47 & 32.01 & 17.70 & 29.01 & 0.19 & 3.58 & 0.00 & 31.43 & 0.19 & 20.00 & 12.14\\
 & Llama-2-7B & 9.67 & 0.77 & 0.77 & 16.83 & 14.99 & 1.06 & 12.19 & 0.77 & 0.00 & 0.00 & 12.09 & 0.58 & 8.29 & 3.34\\
 & Deepseek-7B & 16.34 & 0.77 & 2.71 & 26.50 & 23.99 & 0.58 & 21.57 & 0.19 & 20.02 & 0.00 & 20.21 & 0.10 & 17.47 & 4.69\\
 & TinyLlama-1.1B & 3.97 & 0.29 & 1.64 & 22.82 & 13.44 & 0.87 & 10.64 & 0.48 & 5.61 & 0.10 & 7.83 & 0.10 & 7.19 & 4.11\\
\hline
\multirow{3}{*}{Instr.} & Llama-3-Inst & 16.63 & 0.00 & 0.00 & 13.83 & 23.21 & 3.58 & 34.24 & 0.00 & 0.00 & 0.00 & 20.50 & 0.00 & 15.76 & 2.90\\
 & Qwen2.5-Inst & 27.27 & 0.10 & 2.42 & 7.25 & 42.17 & 88.01 & 39.85 & 58.90 & 0.10 & 0.00 & 42.26 & 28.05 & 25.68 & 30.38\\
 & Mistral-Inst & 11.51 & 63.93 & 0.00 & 9.28 & 26.98 & 82.21 & 25.44 & 15.38 & 0.77 & 2.61 & 23.69 & 5.80 & 14.73 & 29.87\\
\hline
\multirow{8}{*}{Code} & Qwen2.5-Coder & 31.82 & 84.72 & 8.90 & 41.49 & 40.33 & 36.27 & 40.72 & 89.56 & 40.14 & 89.56 & 41.39 & 91.20 & 33.88 & 72.13\\
 & Qwen2.5-Coder-I & 35.78 & 89.94 & 0.48 & 1.45 & 49.61 & 94.97 & 54.26 & 93.33 & 3.19 & 10.45 & 46.62 & 58.41 & 31.66 & 58.09\\
 & CodeLlama-7b & 19.15 & 0.39 & 3.77 & 26.02 & 31.43 & 0.58 & 26.98 & 0.00 & 0.77 & 0.00 & 24.37 & 0.29 & 17.75 & 4.55\\
 & CodeLlama-Inst & 22.82 & 28.72 & 1.26 & 23.31 & 33.75 & 0.77 & 26.69 & 4.26 & 23.02 & 3.19 & 29.40 & 0.29 & 22.82 & 10.09\\
 & Deepseek-Coder & 35.59 & 7.16 & 4.26 & 30.08 & 39.75 & 0.77 & 41.20 & 6.77 & 39.56 & 88.59 & 38.59 & 3.29 & 33.16 & 22.78\\
 & Deepseek-Coder-I & 19.15 & 51.64 & 0.10 & 22.05 & 36.27 & 1.06 & 39.85 & 0.00 & 15.67 & 0.00 & 37.43 & 3.00 & 24.74 & 12.96\\
 & StarCoder2-3B & 22.44 & 0.00 & 9.57 & 45.75 & 32.30 & 0.87 & 30.27 & 0.00 & 2.52 & 0.00 & 31.04 & 0.00 & 21.36 & 7.77\\
 & CodeS-1B & 8.90 & 52.22 & 7.93 & 40.72 & 31.14 & 0.10 & 23.69 & 0.29 & 20.02 & 72.44 & 23.31 & 55.71 & 19.17 & 36.91\\
\hline
\multirow{4}{*}{Chat} & DeepSeek-Chat & 17.31 & 63.64 & 0.29 & 10.83 & 28.72 & 45.16 & 29.59 & 40.62 & 5.03 & 0.87 & 27.95 & 14.12 & 18.15 & 29.21\\
 & Vicuna-7b & 5.22 & 14.22 & 0.39 & 4.35 & 11.03 & 22.73 & 12.28 & 27.76 & 2.42 & 7.16 & 5.51 & 16.54 & 6.14 & 15.46\\
 & Llama-2-Chat & 13.54 & 0.10 & 0.19 & 14.12 & 18.67 & 0.10 & 14.31 & 6.77 & 2.22 & 0.19 & 10.83 & 0.19 & 9.96 & 3.58\\
 & TinyLlama-Chat & 4.74 & 10.93 & 0.39 & 12.77 & 15.57 & 40.72 & 12.09 & 9.57 & 9.28 & 19.92 & 10.64 & 19.92 & 8.79 & 18.97\\
\hline
\multirow{4}{*}{SQL} & SQLCoder-7b & 0.87 & 80.17 & 0.00 & 0.00 & 0.29 & 77.37 & 1.45 & 77.18 & 0.58 & 75.44 & 0.58 & 74.86 & 0.63 & 64.17\\
 & Prem-1B-SQL & 6.19 & 71.66 & 0.39 & 18.18 & 56.67 & 7.74 & 42.75 & 87.04 & 40.43 & 88.69 & 42.65 & 60.06 & 31.51 & 55.56\\
 & CodeS-1B-Spider & 27.85 & 65.96 & 42.55 & 74.57 & 41.20 & 77.85 & 42.36 & 72.82 & 41.20 & 71.95 & 41.10 & 71.95 & 39.38 & 72.52\\
 & CodeS-1B-Bird & 15.28 & 69.54 & 21.66 & 64.31 & 20.89 & 73.31 & 19.05 & 69.15 & 21.18 & 69.54 & 21.86 & 69.15 & 19.99 & 69.17\\
\hline
\end{tabular}
}
\caption{Model performance (EM/EX) across prompt styles on the Spider evaluation set.}
\label{tab:prompt_style_results}
\end{table*}

\textbf{Model families.}
Overall, SQL-specialized models achieve the strongest results, with all models in this group reaching an average execution accuracy above 55\%. Code-oriented models generally follow, but their performance is more sensitive to the prompt style. For example, CodeLlama-7B, its instruction-tuned variant, StarCoder2, and DeepSeek-Coder show strong variations across prompt formats, performing well only under certain representations. The instruction-tuned DeepSeek-Coder variant is more robust with Alpaca-style prompts, suggesting better alignment with its training format. Notably, CodeS achieves high accuracy across most prompt styles despite having only 1B parameters. General-purpose instruction-tuned models, such as Qwen2.5-7B-Instruct and Llama-3-8B-Instruct, underperform their code-tuned counterparts, confirming the importance of code and task-oriented fine-tuning. Fundamental models perform worse overall, while chat-oriented models are the weakest family, with average execution accuracy below 30\%.

\textbf{Model size.}
This study targets reliable Text-to-SQL generation with smaller models. While results vary across families, the 1B-parameter models show clear differences: SQL-specialized 1B models reach average execution accuracy above 55\%, followed by the code-oriented 1B model at approximately 36\%, then chat-oriented models at approximately 18\%, and finally fundamental models at approximately 4\%.  These results support two main observations: (i) SQL-specific fine-tuning is particularly important at small model scales, and (ii) prompt choice remains a major factor in achieving strong performance.

\begin{figure*}[!t]
  \centering
  \includegraphics[
    width=0.7\textwidth,
    trim=0 0 0 180,
    clip
  ]{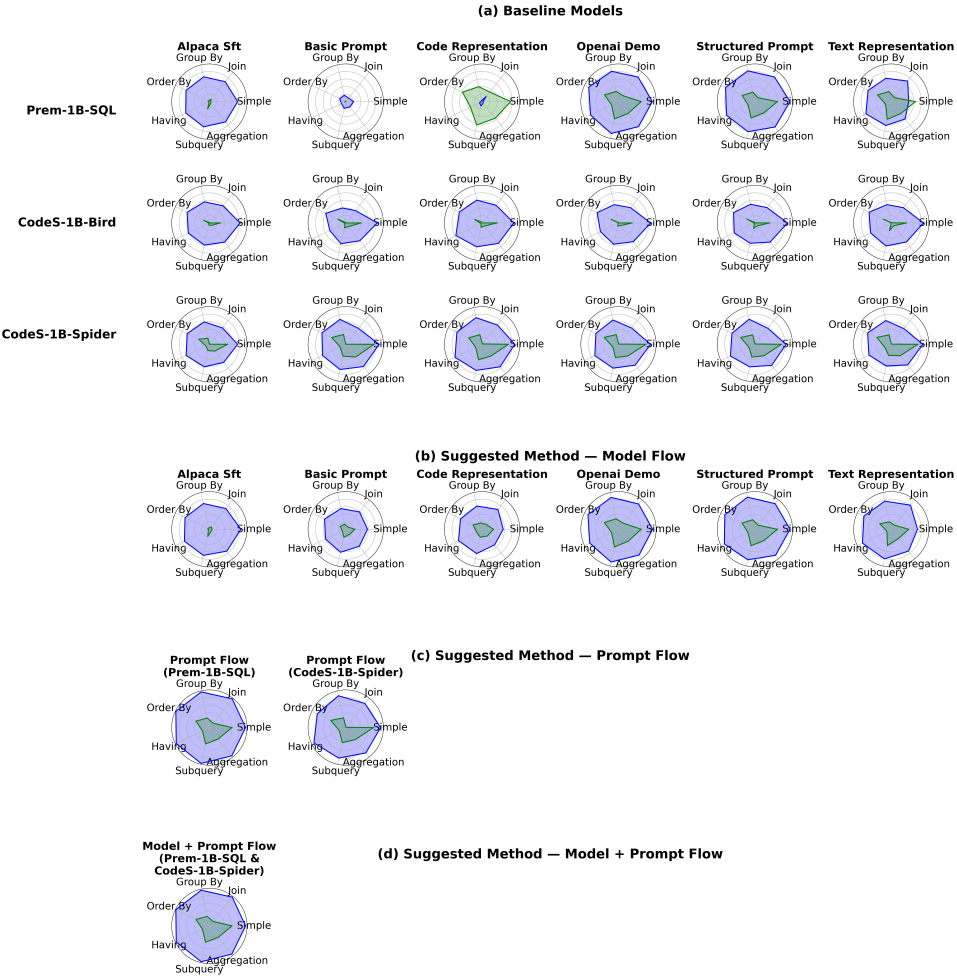}
  \caption{Radar plots of EM (green) and EX (blue) across query categories. (a) Baseline single-pass inference with different prompt styles. (b) Model-Flow workflow. (c) Prompt-Flow workflow. (d) Model-and-Prompt Flow workflow.}
  \label{fig:combined_radar}
\end{figure*}

\textbf{Prompt style.}
Beyond model family, prompt style has a strong impact on performance. The results show that the same model can behave very differently depending on how the schema and question are presented. To analyze this effect, we compare prompt styles across SQL query categories for the 1B SQL-oriented models, which are the main focus of this work in resource-constrained settings. In Fig.~\ref{fig:combined_radar}(a), the green polygon represents EM and the blue polygon represents EX, while each radar axis corresponds to a query category. As shown in Fig.~\ref{fig:combined_radar}(a), performance varies not only by prompt style but also by query type. In particular, the code representation prompt performs better on multi-table queries, likely because it provides richer schema information. Overall, no single prompt style is uniformly optimal across models and query categories, making prompt selection an important component of a Text-to-SQL system.

Based on this observation, we evaluate the three proposed workflows introduced in Section~\ref{sec:workflows}. The goal is to improve robustness by adapting the inference strategy after a failed attempt, either by changing the prompt style, switching to another model, or combining both. Table~\ref{tab:workflow_results} summarizes the accuracy of these workflows.

\begin{table}[h!]
\centering
\tiny
\caption{Workflow Accuracy Comparison (EX / EM)}
\resizebox{\columnwidth}{!}{%
\begin{tabular}{l c c c}
\hline
Algorithm & Repr & EX & EM \\
\hline
Prompt workflow (Prem-1B-SQL) & -- & \textbf{97.20} & 41.30 \\
Prompt workflow (CodeS-1B-Spider) & -- & 87.81 & 43.13 \\
\hline
Model and prompt workflow & -- & \textbf{97.20} & 41.30 \\
\hline
\multirow{6}{*}{Model workflow} & Alpaca Sft & 76.69 & 7.83 \\
 & Basic Prompt & 58.99 & 16.34 \\
 & Code Repr. & 62.96 & 23.89 \\
 & OpenAI Demo & 88.78 & 43.13 \\
 & Struct. Prompt & 90.33 & 40.62 \\
 & Text Repr. & 77.27 & 33.46 \\
\hline
\end{tabular}
}
\label{tab:workflow_results}
\end{table}
The prompt-flow workflow is evaluated with both \texttt{Prem-1B-SQL} and \texttt{CodeS-1B-Spider}. In both cases, it improves execution accuracy compared to any single prompt style, showing that varying $P$ after failure can substantially reduce errors. By contrast, model-flow generally provides smaller gains, suggesting that changing $M$ alone is not sufficient when prompt formatting is a major source of failure. Finally, the model-and-prompt workflow reaches the same execution accuracy as the best prompt-flow configuration, but requires more inference calls, making it less suitable for resource-constrained settings.

Overall, prompt-flow offers the best accuracy--cost trade-off in our setup: it improves robustness while keeping the number of additional inference calls moderate. When maximum accuracy is the main objective, model-and-prompt flow can further reduce failures, but with higher computation. Thus, all workflows trade additional inference cost for higher reliability, and the best choice depends on the deployment constraints.

Fig.~\ref{fig:combined_radar} compares the baseline and the three proposed workflows across query categories. Compared to the baseline in Fig.~\ref{fig:combined_radar}(a), the workflows generally expand the EX profile (blue), indicating fewer execution failures. Model-Flow in Fig.~\ref{fig:combined_radar}(b) gives moderate improvements, while Prompt-Flow in Fig.~\ref{fig:combined_radar}(c) provides more consistent gains, showing that adapting $P$ after failure is effective for small Text-to-SQL models. Finally, Model-and-Prompt Flow in Fig.~\ref{fig:combined_radar}(d) achieves the strongest overall profile, but at the cost of more inference calls, consistent with Table~\ref{tab:workflow_results}.

\section{Conclusions}\label{sec:conclusion}
This work investigated a range of LLM families for the Text-to-SQL task. The results show that SQL-specific fine-tuning consistently outperforms other model families, and that this gap becomes more pronounced at smaller model scales (around 1B parameters).

A key limitation observed in practice is the unstable behavior of a fixed model--prompt pair across different query categories, which makes single-pass inference brittle. To address this, we proposed three workflows and evaluated them against a baseline single-model/single-prompt setting. Overall, Prompt-Flow provides the best robustness under resource constraints, while Model-and-Prompt Flow achieves the strongest performance when accuracy is prioritized, at the cost of higher computation.

Future work should focus on learning or predicting an efficient ordering of model and prompt selection based on the input question, with the goal of reducing inference cost while maintaining high accuracy.

\bibliographystyle{IEEEtran}
\bibliography{references}

\end{document}